\documentclass[10pt,twocolumn,letterpaper]{article}

\usepackage{cvpr}

\definecolor{revised}{RGB}{120,69,230}

\definecolor{cvprblue}{rgb}{0.21,0.49,0.74}
\usepackage[pagebackref,breaklinks,colorlinks,allcolors=cvprblue]{hyperref}

\title{Rethinking Normalization Strategies and Convolutional Kernels \\ for Multimodal Image Fusion}

\author{Dan He$^{1, \thanks{These authors contributed equally}}$\quad Guofen Wang$^{2,\footnotemark[1]}$\quad Weisheng Li$^{1,\thanks{Corresponding author}}$\quad Yucheng Shu$^{1,\footnotemark[2]}$\quad Wenbo Li$^{1}$\quad
\\ Lijian Yang$^{1}$\quad Yuping Huang$^{1}$\quad Feiyan Li$^{1}$\quad \\
$^1$ Chongqing University of Posts and Telecommunications \quad
$^2$ Chongqing Normal University \\
{\tt\small d230201011@stu.cqupt.edu.cn}
}

\begin{document}
\maketitle
\begin{abstract}
Multimodal image fusion (MMIF) integrates information from different modalities to obtain a comprehensive image, aiding downstream tasks. 
However, existing research focuses on complementary information fusion and training strategies, overlooking the critical role of underlying architectural components like normalization and convolution kernels. 
We reevaluate the UNet architecture for end-to-end MMIF, identifying that widely used batch normalization limits performance by smoothing crucial sparse features. 
To address this, we propose a hybrid of instance and group normalization to maintain sample independence and reinforce intrinsic feature correlations. Crucially, this strategy facilitates richer feature maps, enabling large kernel convolution to fully leverage its receptive field, enhancing detail preservation. 
Furthermore, the proposed multi-path adaptive fusion module dynamically calibrates features from varying scales and receptive fields, ensuring effective information transfer. 
Our method achieves SOTA objective performance on MSRS, M$^3$FD, TNO, and Harvard datasets, producing visually clearer salient objects and lesion areas. Notably, it improves MSRS segmentation mIoU by 8.1\% over the infrared image.
This performance stems from a synergistic design of normalization and convolution kernels, which preserves critical sparse features.
The code is available at {https://github.com/HeDan-11/LKC-FUNet}.
\end{abstract}    
\section{Introduction}
\label{sec:intro}
Image fusion technology fully integrates the complementary information of different modalities to generate a comprehensive representation of the image.
It is widely used for scene information enhancement or recovery.
In addition, downstream tasks such as object detection~\cite{liu2022target} and semantic segmentation~\cite{Ming2025SSDFusionAS, Tang2022PIAFusionAP} can also benefit from sharper scene and object representations in the fused images.
Fusion tasks include infrared and visible image fusion (IVIF)~\cite{Li2023LRRNetAN,wang2024general}, medical image fusion (MIF)~\cite{Liu2024MMNetAM,Tang2024FATFusionAF}, multi-exposure image fusion, and so on, where MIF and IVIF are the two most representative unsupervised fusion scenarios.
Specifically, IVIF overcomes the limitations of poor lighting in visible images and the low resolution of infrared images~\cite{Zhao_EMMA}.
Similarly, MIF generates images that integrate anatomical and metabolic information, which is crucial for improving diagnostic reliability~\cite{He2024MMIFINetMM}.

Traditional frameworks are largely based on multiscale transformations, sparse representations, and subspaces. Their drawback is that they are constrained by reliance on limited fusion rule selection and the complexity of manual design. 
In contrast, deep learning-based methods adaptively learn complex patterns from data, significantly improving fusion efficiency and performance.
Common encoder-decoder models use convolutional neural networks (CNNs)~\cite{Zhang2021SDNetAV,Zhang2020IFCNNAG}, Transformers~\cite{guo2025sam, Zhu2024TaskCustomizedMO}, and Mamba~\cite{Liu2025SSEFusionSS} to extract features and reconstruct images.
Generative Adversarial Networks (GANs)-based methods~\cite{liu2022target,Fu2021DSAGANAG} preserve texture details and highlight salient features through adversarial training. However, the inherent instability of GAN training remains a bottleneck, often leading to discontinuous edges or artifacts. 
Although deep learning methods have achieved satisfactory performance, several issues persist. 

\begin{figure}[!t]
\centering
    \includegraphics[width=1.0\linewidth]{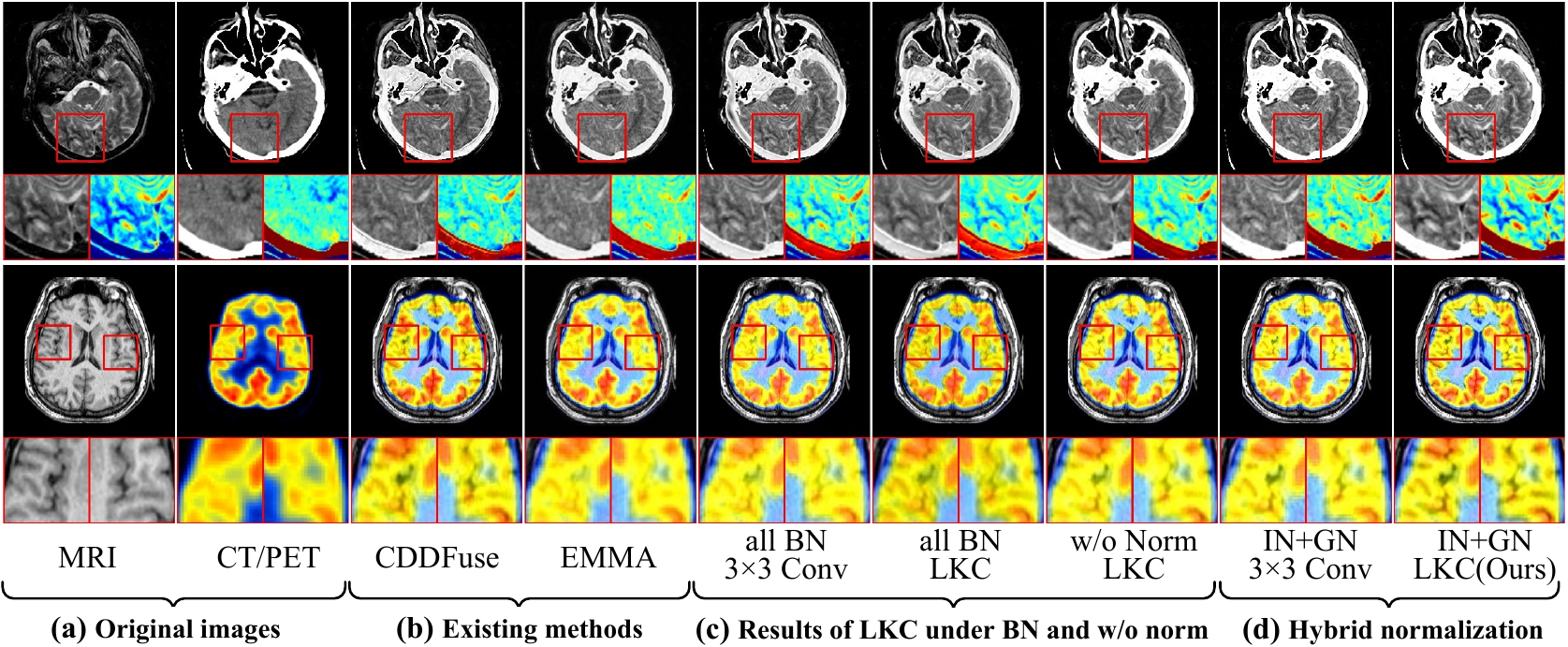}
    \caption{Impact of existing methods, diverse normalization strategies, and large kernel convolution (LKC) on fusion performance.}
    \label{fig:Fig1}
    \vspace{-0.3cm}
\end{figure}

As shown in \Cref{fig:Fig1}(b), some general-purpose fusion methods, such as EMMA~\cite{Zhao_EMMA} and MMDRFuse~\cite{Deng2024MMDRFuseDM}, are successful in IVIF but lose significant detail when applied to MIF. They overlook that medical images possess substantially higher statistical values, such as average gradient (AG) and spatial frequency (SF), than natural images. 
Other methods like SDNet~\cite{Zhang2021SDNetAV} and CDDFuse~\cite{Zhao2022CDDFuseCD} attempt to address this by training separate parameters for each task. However, they neglect inter-sample interference and the degradation of crucial features within the highly sparse data distributions of MIF, causing details to be weakened. 
This oversight persists even in recent state-of-the-art (SOTA) methods like ECFusion~\cite{wei2025ecfusion}, FusionMamba~\cite{xie2024fusionmamba}, and DM-FNet~\cite{He2025DMFNetUM}. These works primarily focus on improving complementary information fusion and training strategies, yet they overlook the importance of underlying components-like normalization and convolution kernels-for sparse feature extraction.

Specifically, many fusion frameworks~\cite{Zhang2020IFCNNAG, Fu2021DSAGANAG} derived from high-level vision tasks widely adopt batch normalization (BN). BN normalizes features across an entire batch, a process that ignores sample independence and causes data smoothing. This may have a lesser impact on IVIF, which emphasizes structure. For medical images, however, this inter-sample influence creates a conflict. These images are characterized by sparse distributions and require strict detail retention, making it difficult to preserve both high-intensity regions and fine details simultaneously. Although some methods~\cite{Deng2024MMDRFuseDM,Liu2023SegMIF} forgo normalization to maintain sample independence, their performance gains are limited. This is because they do not account for inherent image properties or feature relationships. This insight motivates us to rethink the fundamental components of fusion networks.

Furthermore, large kernel convolution (LKC) can capture spatial information over a larger receptive field and is highly effective at preserving image structure. Its exploration in the fusion domain is limited, likely due to performance bottlenecks. 
As depicted in \Cref{fig:Fig1}(c), when BN is used, inter-sample interference smooths data and reduces effective features. 
Consequently, the larger receptive field of LKC provides limited or even detrimental effects on detail preservation. Omitting normalization fails to provide better image characteristics and struggles to raise the upper bound of fusion performance. 
Notably, Transformers and Mamba can serve as alternatives for obtaining a larger receptive field. However, they are more computationally expensive and require large training datasets, which are often unavailable for medical tasks. Additionally, while Transformers emphasize global attention modeling, they are not as good at capturing local information, resulting in smoother features that tend to lose high-frequency details~\cite{si2022inception}.
Finally, in standard UNet architectures, simple skip connections do not consider the relative importance of feature maps from different paths. This may cause crucial features to be overlooked.

To address the above issues, we employ a mixture of IN and GN to fully consider sample independence, image properties, and intrinsic feature connections. This strategy enhances the generation of rich feature maps while more effectively preserving the distinctive attributes of source images, as illustrated in \Cref{fig:Fig1}(d). It also elevates the upper bound of fusion performance. 
At this point, applying LKC further enhances detail retention. 
Finally, the feature maps in the input decoder are recalibrated by combining spatial, channel attention, and bidirectional interactions. 

In summary, we focus on the synergy between normalization and convolution on the unique statistical properties of multimodal images, thereby maximizing the preservation of modality-specific and sparse details. Based on this, our method guarantees outstanding performance in MIF and remains fully applicable to IVIF. 
Our contributions are summarized as follows:
\begin{itemize}
\item A UNet with LKC is proposed to achieve multimodal image fusion (MMIF), namely LKC-FUNet, including IVIF and MIF.
\item Rethink the impact of normalization and LKC on image fusion. Verify the inappropriateness of BN in MMIF. Mixing IN and GN preserves image properties and is well-suited for highly sparsely distributed fusion tasks. Under the above strategy, LKC enlarges the "effective receptive field" to protect detailed information better and significantly improve fusion performance. 
\item A multipath adaptive fusion module is designed for feature fusion across different receptive fields and scales. Spatial-channel dual-attention feature maps, bidirectional interactions, and recalibration provide more comprehensive inputs to the decoder.
\item Our method significantly improves multiple metrics in both tasks and achieves breakthroughs in MIF visualization. It has also been shown to facilitate downstream multimodal semantic segmentation.
\end{itemize}
\section{Related work}
\label{sec:related}
\subsection{Multimodal image fusion} 
Deep learning methods in MMIF can be categorized into four groups: encoder-decoder models, adversarial models, task-driven models, and others. 
\begin{table*}[!ht]
\footnotesize
    \centering
    \caption{Comparative analysis of SOTA fusion methods}
    \label{tab:sota}
    \begin{tabular}{llllll}
    \hline
        Method & Year & Architecture & Norm & Main Focus & Scenarios \\ \hline
        DSAG~\cite{Fu2021DSAGANAG} & 2021 & GAN & BN & Adversarial training with dual-stream attention. & MIF \\
        FATF~\cite{Tang2024FATFusionAF} & 2024 & Transformer & BN & Modality-specific feature extraction and interaction. & MIF \\ 
        DM-FN~\cite{He2025DMFNetUM} & 2025 & UNet & GN & Two-stage training (diffusion reconstruction + fusion). & MIF \\ 
        LRRN~\cite{Li2023LRRNetAN} & 2023 & CNN & \(\times\) & Model interpretability and model efficiency. & IVIF \\
        DCIN~\cite{wang2024general} & 2024 & Invertible Network & \(\times\) & Feature extraction process with lossless information. & IVIF \\ 
        MaeF~\cite{Li2024MaeFuseTO} & 2025 & Autoencoder & BN & Using pre-trained omni-features (low \& high level) from MAE. & IVIF \\ 
        EMMA~\cite{Zhao_EMMA} & 2024 & Restormer-CNN & LN & Self-supervised training via equivariant imaging prior. & IVIF, MIF \\
        MMDR~\cite{Deng2024MMDRFuseDM} & 2024 & CNN & \(\times\) & Model efficiency (knowledge distillation). & IVIF, MIF \\ 
        MMAE~\cite{Wang2024MMAEAU} & 2024 & Autoencoder & LN & Information filtering via mask attention mechanism. & IVIF, MIF \\ 
        MLFu~\cite{10856398_MLFuse} & 2025 & CNN & BN & Deconstructing fusion into three sub-tasks (reinforcing, edge-guiding). & IVIF, MIF \\ 
        VDMU~\cite{10794610_VDMFusion} & 2025 & Diffusion Model & GN & Integrating fusion into diffusion sampling as a weighted average process. & IVIF, MIF \\
        Our & 2025 & UNet & IN+GN & Rethinking underlying components (normalization and convolution). & IVIF, MIF \\ \hline
    \end{tabular}
\end{table*}

Early encoder-decoder~\cite{Zhang2020IFCNNAG} models used pre-trained autoencoders for feature extraction and reconstruction, with an emphasis on designing fusion rules. Later, end-to-end models were developed to overcome the limitations of manual rules. These models primarily combine CNNs~\cite{10856398_MLFuse}  and Transformers~\cite{Tang2024FATFusionAF} to enrich feature representations and facilitate information interaction between modalities. 
GAN is widely applied to address the challenge of data scarcity~\cite{alzubaidi2023survey}, particularly in unsupervised tasks like image fusion, where supervised ground-truth labels are unavailable. In such tasks, the adversarial mechanism provides a powerful implicit supervisory signal. For instance, 
DSAGAN~\cite{Fu2021DSAGANAG} and TarDAL~\cite{liu2022target} model image fusion as a competition between a generator and a discriminator, preserving texture details and highlighting salient targets through adversarial learning. 
However, the instability of GAN training remains a long-standing challenge.

In recent years, research has shifted towards improving the adaptability of fused images for downstream tasks. CDDFuse~\cite{Zhao2022CDDFuseCD} and EMMA~\cite{Zhao_EMMA} measure fusion performance based on object detection and segmentation accuracy. Other studies~\cite{Liu2023SegMIF, liu2025hsenet} guide the fusion process using semantic information from downstream tasks. Meanwhile, diffusion-based fusion methods~\cite{Li2023FusionDiffMI} transform the fusion problem into a probabilistic generation task and embed inference mechanisms within iterative diffusion processes, thereby demonstrating enhanced robustness in image fusion tasks. DM-FNet~\cite{He2025DMFNetUM} employs diffusion processes to train its encoder-decoder architecture, significantly enhancing the model's feature representation capabilities. This improvement stems from the self-supervised denoising~\cite{zhang2023unleashing} process in the first stage.

As summarized in \Cref{tab:sota}, existing research has largely focused on network architecture design, semantic information capture, and training strategy optimization. However, the intrinsic impact of fundamental modules like normalization on image fusion remains underexplored. We address this gap and show that a hybrid IN+GN normalization effectively aligns the statistical distributions of different modalities, which is essential for fully exploiting the feature extraction capability of LKC.

\subsection{Normalization}
Normalization techniques stabilize and accelerate model training by unifying data scales, but their application and systematic study in image fusion are insufficient. Most fusion models either adopt network architectures designed for other vision tasks or omit normalization entirely.

A key challenge in image fusion is handling the significant statistical heterogeneity between source image pairs. This makes the choice of normalization strategy critical, as single strategies often have inherent limitations. 
BN~\cite{ioffe2015batch}, while accelerating convergence, introduces statistical coupling across samples, creating a "smoothing" effect that harms the preservation of modality-specific information.
LN normalizes across all feature channels of a single sample. In a channel-concatenated input scenario, this forces the statistics of different modalities to unify, thereby weakening the unique distributional properties of each modality.
IN~\cite{Ulyanov2016IN} processes each channel independently, which helps to suppress style differences and highlight structural details. However, improper use can excessively amplify distributional differences between modalities.
GN~\cite{Wu2018GN} balances channel locality with stability, but its adaptability to modal differences is still limited when used alone. 
BCN adaptively selects between BN and LN, but its performance remains constrained by BN's smoothing effect.

Clearly, a single normalization strategy struggles to balance the need to preserve modality commonalities and extract unique characteristics. Therefore, designing a hybrid normalization strategy for MMIF that accommodates different statistical distributions is key to unlocking a model's full performance potential.

\begin{figure*}[!t]
\centering
    \includegraphics[width=0.85\linewidth]{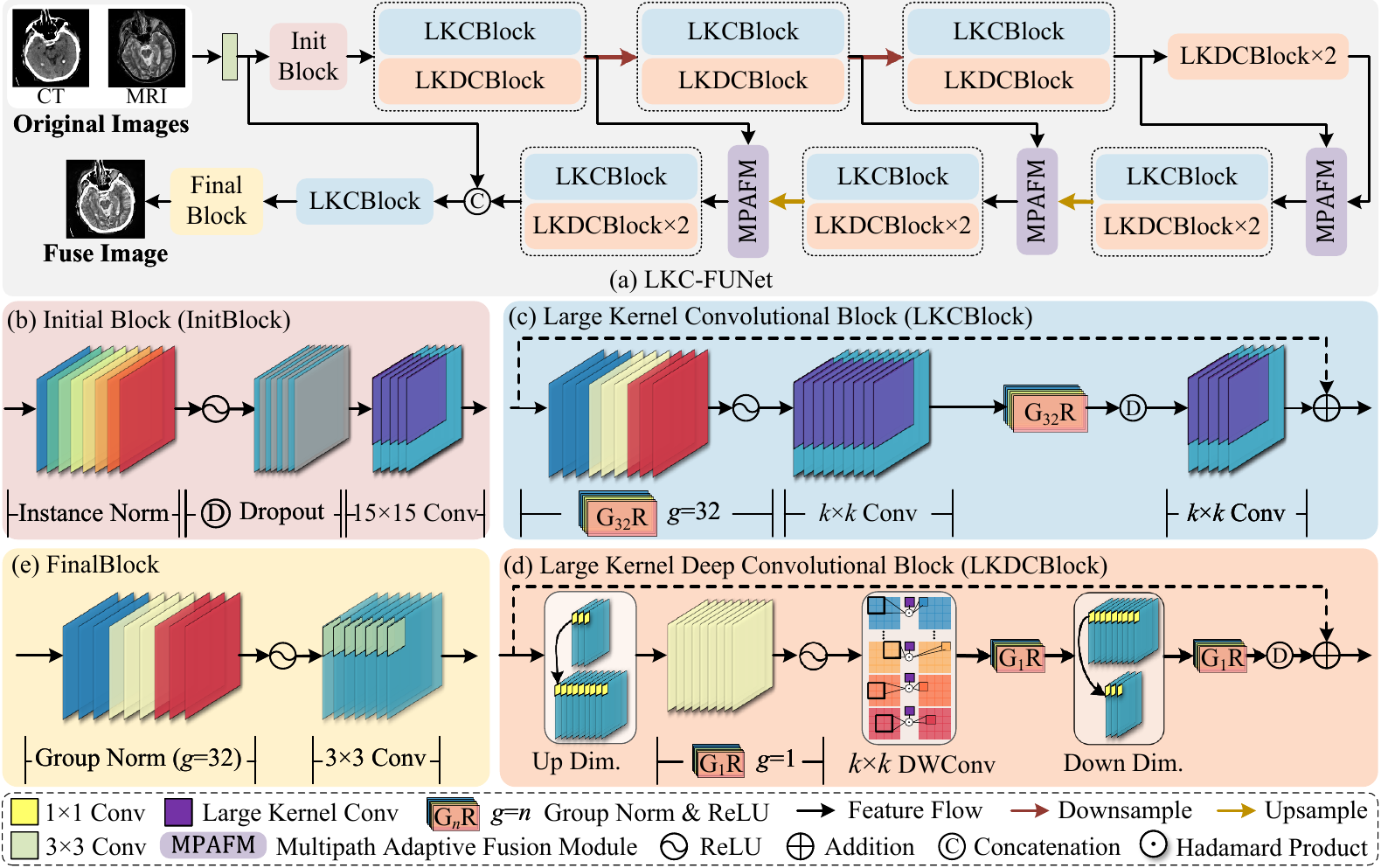}
    \caption{Overall architecture of LKC-FUNet.}
    \label{fig:overall_network}
    \vspace{-0.2cm}
\end{figure*}
\subsection{Large kernel convolution}
LKC is widely used in various vision tasks to acquire richer contextual features by expanding the receptive field. For instance, LR-Net~\cite{hu2019LRNet} and ConvNeXt~\cite{liu2022convnet} benefit from a 7$\times$7 kernel. Howerver, LR-Net's performance degrades at a 9×9 size, indicating that performance bottlenecks and parameter spikes limit the kernel size. To address this, RepLKNet~\cite{ding2022RepLKNet} achieves superior performance by reparameterizing a 31$\times$31 depthwise convolution, while SLaK extends the kernel size to 51$\times$51 through kernel decomposition and sparse grouping.

Although LKC has been proven effective in classification and segmentation, its integration in image fusion has not been systematically studied due to performance bottlenecks caused by certain normalization methods. Traditional small-kernel convolutions tend to introduce excessive smoothing when stacked in multiple layers, leading to detail loss. LKC can theoretically mitigate this by covering a larger area in a single layer, thus requiring less network depth. However, standard normalization methods like BN smooth the feature maps, diminishing feature richness and preventing LKC's large receptive field from being effective.

\section{Method}
\label{sec:method}
\subsection{Overview}
The proposed method is a UNet architecture, as shown in \Cref{fig:overall_network}. We splice the source image pairs into the input model to completely avoid the design of inter-modal fusion rules.
The basic modules are the initial block, the Large kernel convolutional block, and the Large kernel deep convolutional block. Their design follows a core principle: preserving independent image properties with IN in the shallow layers, strengthening feature correlations with GN in the deep layers, and expanding the effective receptive field throughout the network using LKC. Finally, a multipath adaptive fusion module recalibrates the decoder inputs.

\subsection{Basic modules}
\textbf{Initial block (InitBlock).}
Our InitBlock combines IN and LKC with 15$\times$15.
IN normalizes each multi-modal input image independently. This is crucial for reducing the statistical gap between modalities that have very different properties, while simultaneously preserving fine-grained details. In parallel, the 15x15 LKC captures a wide receptive field, which strengthens the overall image structure and maintains contrast.
Considering the computational efficiency and the intrinsic relationship between features, GN is adopted for the other modules.

\begin{figure*}[!t]
\centering
    \includegraphics[width=0.7\linewidth]{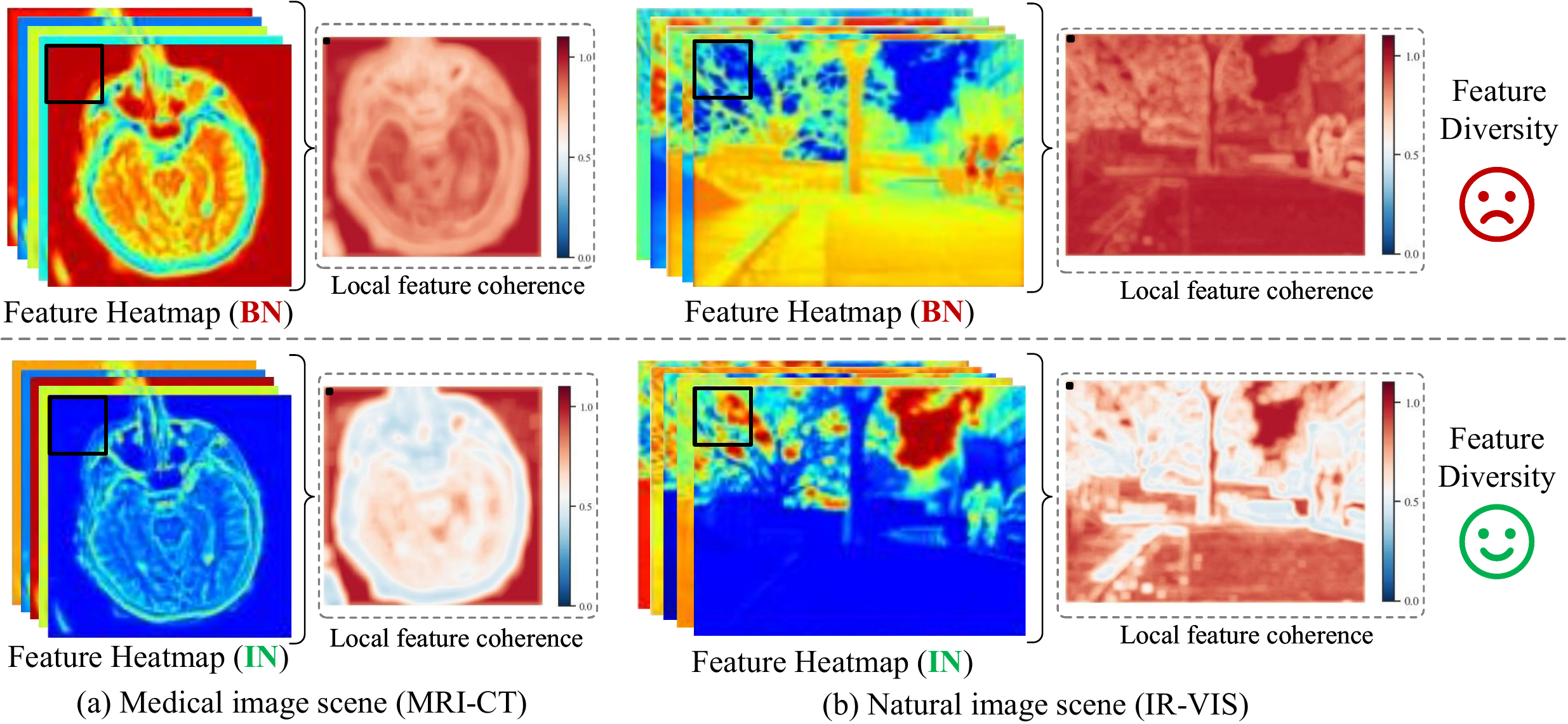}
    \caption{Analysis of feature redundancy for different normalization strategies in the InitBlock. Higher values indicate more redundant features in the corresponding patches.}
    \label{fig:coh_fea}
\vspace{-0.5cm}
\end{figure*}

To validate the impact of different normalization strategies on feature redundancy in the InitBlock, we compute the inter-channel correlation within local neighborhoods of the feature map. Higher values indicate more feature redundancy.
As illustrated in \Cref{fig:coh_fea}, the background regions under both BN and IN exhibit high consistency. However, in areas with complex structures and fine details (e.g., building edges, foliage), the feature map generated under IN shows significantly lower redundancy compared to BN. Therefore, IN combined with LKC generates a richer feature representation, which is crucial for preserving the structural integrity and detail.

\textbf{Large kernel deep convolutional block (LKCBlock).} 
The LKCBlock comprises two sequential modules, each containing GN, ReLU, an optional dropout, and a convolutional layer. A final residual connection ensures stable gradient flow and preserves essential information.
Given the $64\times64$ input image size during training, the convolution kernel sizes at different stages are set to 15, 11, 5, and 5. In the high-resolution stage, the $15\times15$ kernel captures a wide receptive field, allowing the model to grasp global structural information and long-range dependencies from the outset. As the feature map resolution decreases, the kernel sizes are progressively reduced to $11\times11$ and $5\times5$. This strategy maintains a large receptive field relative to the feature map size at each stage, while also reducing computational overhead.

\textbf{Large kernel deep convolutional block (LKDCBlock).}
Considering the model's efficiency, we use LKDCBlock~\cite{ding2022RepLKNet} combined with LKCBlock.
It is worth noting that the grouping of GN in LKDCBlock is set to 1.
This change is based on the property of deepwise convolution, in which convolution kernels independently operate on corresponding feature maps. At this point, grouping the GN may result in weak connections between features and amplification of certain features.
Moreover, the encoder focuses on feature extraction, and using too many LKDCBlocks can significantly increase computational complexity. The decoder requires higher modeling capacity to balance local details and global contrast preservation during feature recovery. Increasing the number of LKDCBlock helps refine the output image quality, so we added one more block to the decoder.

\subsection{Multipath adaptive fusion module (MPAFM)}
\begin{figure}[!ht]
\centering
    \includegraphics[width=1.0\linewidth]{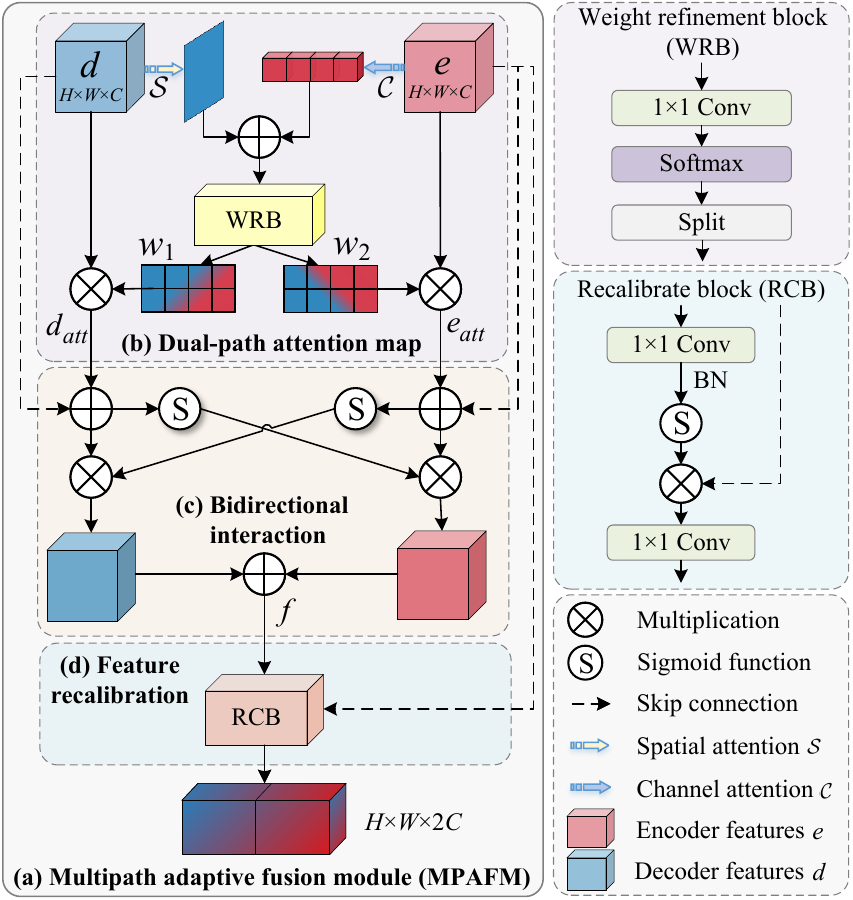}
    \caption{Network of MPAFM.}
    \label{fig:MPAFM}
\vspace{-0.5cm}
\end{figure}

Simple skip connections in UNet do not consider the relative importance of feature maps in different paths during the fusion process, which may result in some critical features being overlooked~\cite{Azad2022MedicalIS, Kolahi2024MSA2NetMA}. Therefore, this section proposes the MPAFM to process the feature maps of the corresponding layers of the encoder and decoder, as shown in \Cref{fig:MPAFM}. The aim is to suppress unimportant features to minimize artifacts while progressively capturing detailed and structural information for the image at different scales and receptive fields.
It operates through three distinct stages: Dual-path attention feature generation, bidirectional interaction, and feature recalibration.

\textbf{Dual-path attention feature generation.}
This stage aims to generate two enhanced feature maps, each selectively focused on the most critical information from its respective path.
First, we use channel attention ($\mathcal{C}$) and spatial attention ($\mathcal{S}$) to process the encoder-decoder features ($e$ and $d$), generating a pre-attention map $A_{pre}$. This initial fusion captures both "what" (channel) and "where" (spatial) is important. BN ($\mathcal{B}$) is then used to stabilize the feature distribution, which helps the subsequent layer learn more stable spatial weights. It is defined as:
\begin{equation}
A_{pre} = \mathcal{B}(\mathcal{C}(e) + \mathcal{S}(d))
\label{eq1}
\end{equation}
Next, $A_{pre}$ is refined into $w_e$ and $w_d$ using the weight refinement block $\mathcal{W}$. $\mathcal{W}$ contains 1$1\times$1 selection convolution $\varphi_{1\times1}$, softmax function  $\sigma$, and split operations, as shown in \Cref{eq2}.
This competitive allocation forces the model to decide at each pixel whether to prioritize information from the encoder or the decoder.
\begin{equation}
w_1, w_2 = Split(\sigma(\varphi_{1\times1}(A_{pre})))
\label{eq2}
\end{equation}
Finally, $w_1$ and $w_2$ are applied to their respective encode-decoder features ($e$ and $d$), to produce the final attention-enhanced features, as shown in \Cref{eq3}.
\begin{equation}
e_{att} = w_1 \odot e, \quad d_{att} = w_2 \odot d
\label{eq3}
\end{equation}
where $\odot$ denotes the dot product operation.

\textbf{Bidirectional Interaction.}
This stage uses a cross-gating mechanism to facilitate a mutual exchange and enhancement of features between the two paths. First, a residual connection preserves the original information while applying the learned focus, preventing feature degradation, as shown in \Cref{eq4}.
\begin{equation}
\tilde{e} = e_{att} + e, \quad \tilde{d} = d_{att} + d
\label{eq4}
\end{equation}
Subsequently, a gating signal is generated via a sigmoid function to modulate the information from the opposing path, as shown in \Cref{eq5}. This allows features from different scales and receptive fields to interact dynamically.
\begin{equation}
f = \sigma(\tilde{e}) \odot \tilde{d} + \sigma(\tilde{d}) \odot \tilde{e}
\label{eq5}
\end{equation}

\textbf{Feature Recalibration}. 
The final stage uses the rich contextual information from the interaction to recalibrate the original encoder features. The interacted feature map is processed by a 1$1\times$1 Conv, BN, and sigmoid function, as shown in \Cref{eq6}. This step distills the fused context into an efficient pixel-wise attention map $A$. 
\begin{equation}
A = \sigma(\mathcal{B}(\varphi_{1\times1}(f)))
\label{eq6}
\end{equation}
This map is then used to recalibrate the encoder feature $e$ with the comprehensive information learned during the interaction, enhancing the encoder features at a fine-grained level. A final 1$1\times$1 Conv is used for dimension, as shown in \Cref{eq7}.
\begin{equation}
X = \varphi_{1\times1}(A \odot e)
\label{eq7}
\end{equation}

\subsection{Loss function}
High-quality image fusion has three key objectives: preserving structural integrity, retaining salient intensity, and maintaining sharp details. A single loss function struggles to address all these goals simultaneously. We therefore use a composite loss function that combines three distinct terms for structure, intensity, and gradient. This design achieves a balanced optimization for the final fused image. The total loss is defined as follows:
\begin{equation}
   {\mathcal{L}}_{total}={\mathcal{L}}_{ssim}+{\mathcal{L}}_{int}+{\mathcal{L}}_{grad}
\end{equation}

\textbf{Structural Similarity Loss.} This loss aligns with human visual perception by comparing the structural content of images rather than absolute pixel differences. It encourages the network to generate a result that is structurally consistent with the input images, calculated as follows:
\begin{equation}
   {\mathcal{L}}_{ssim}=1-0.5\cdot(SSIM(I_{A}^{Y},I_{F}^{Y})+SSIM(I_{B}^{Y},I_{F}^{Y}))
\end{equation}
where SSIM is the structural similarity. $I_{A}$ and $I_{B}$ are the source images, $I_{F}$ is the fused image. $I_{A}^{Y}$, $I_{B}^{Y}$, $I_{F}^{Y}$ are the respective luminance channels.

\textbf{Intensity Loss.} This term constrains the pixel values of the fused image to be close to the most significant intensity values from the sources at each location. This ensures the preservation of salient intensity features, such as prominent thermal targets in an infrared image. It is defined as:
\begin{equation}
   {\mathcal{L}}_{int}={\frac{1}{HW}}\cdot\left\|I_{F}^{Y}-\max(I_{A}^{Y},I_{B}^{Y})\right\|_{1}
\end{equation}
where  $\|\cdot\|_1$ denotes the $L_1$ norm and $max(\cdot)$ represents the element-wise maximum operation.

\textbf{Gradient loss.} Operating in the gradient domain, this term directly penalizes any loss of high-frequency details. It counteracts the smoothing tendency of neural networks and ensures the fused image has sharp edges and fine textures. It is defined as: 
\begin{equation}
   {\mathcal{L}}_{grad}={\frac{1}{HW}}\cdot\left\|{|{\boldsymbol{\nabla}}I_{F}^{Y}|-\max{\big(}|{\boldsymbol{\nabla}}I_{A}^{Y}|,|{\boldsymbol{\nabla}}I_{B}^{Y}|{\big)}}\right\|_{1}
\end{equation}
where $\nabla$ indicates the gradient operator used for texture information measurement within an image.
\section{Experiment}
\label{sec:relults}
\begin{table*}[!t]
\footnotesize
\setlength\tabcolsep{5pt}
\renewcommand\arraystretch{1.05}
\caption{Ablation experiment results in the test set of MIF. Red: best, blue: second-best.}
\label{tab:xrsy}
\vspace{-0.2cm}
\begin{center}
    \begin{tabular}{l|lcc|cccccc}
    \hline
    \textbf{No.} & \textbf{Norm} & \textbf{LKC} & \textbf{MPAFM} & \multicolumn{1}{c}{SD} & \multicolumn{1}{c}{AG} & \multicolumn{1}{c}{SF} & \multicolumn{1}{c}{SCD} & \multicolumn{1}{c}{VIFF} & \multicolumn{1}{c}{SSIM} \\ \hline
    a & BN & $\times$ & $\checkmark$ & 76.553 & 8.146 & 28.280 & 1.578 & 0.548 & 0.736 \\
    b & BN & $\checkmark$ & $\checkmark$ & 73.995 & 7.282 & 24.928 & 1.485 & 0.517 & \textbf{\textcolor{red}{0.748}} \\
    c & $\times$ & $\checkmark$ & $\checkmark$ & 78.414 & 7.621 & 26.807 & 1.584 & 0.543 & \textbf{\textcolor{blue}{0.747}} \\
    d & LN & $\checkmark$ & $\checkmark$ & 76.196 & 7.256 & 24.116 & \textbf{\textcolor{blue}{1.620}} & 0.537 & 0.737 \\
    e & BCN & $\checkmark$ & $\checkmark$ & 76.080 & 6.646 & 22.446 & 1.572 & 0.530 & 0.740 \\
    f & IN  & $\checkmark$ & $\checkmark$ & 74.086 & \textbf{\textcolor{blue}{9.415}} & \textbf{\textcolor{red}{38.771}} & 0.700 & 0.462 & 0.636 \\
    g & GN & $\checkmark$ & $\checkmark$ & 80.503 & 8.406 & 28.969 & \textbf{\textcolor{red}{1.640}} & \textbf{\textcolor{blue}{0.580}} & 0.720 \\
    h & BN+GN & $\checkmark$ & $\checkmark$ & 78.506 & 8.170 & 26.516 & 1.609 & 0.552 & 0.746 \\ \hline
    i & IN+GN & $\times$ & $\checkmark$ & 79.999 & 8.520 & 30.974 & 1.591 & 0.571 & 0.735 \\
    j & IN+GN & $\checkmark$ & $\times$ & \textbf{\textcolor{blue}{81.203}} & 8.908 & 32.373 & 1.535 & 0.569 & 0.725 \\
    k & {IN+GN} & \textbf{$\checkmark$} & \textbf{$\checkmark$} & \textbf{\textcolor{red}{83.422}} & \textbf{\textcolor{red}{9.716}} & \textbf{\textcolor{blue}{35.381}} & 1.618 & \textbf{\textcolor{red}{0.586}} & 0.722 \\ \hline
    \end{tabular}
\end{center}
\vspace{-0.5cm}
\end{table*}
\subsection{Medical image fusion}
\subsubsection{Step}
\textbf{Datasets.} We selected three medical image combinations from the Harvard Medical School website \footnote[1]{Harvard Medical website: http://www.med.harvard.edu/AANLIB/home.html} for MIF experiments, including MRI-CT, MRI-PET, and MRI-single-photon
emission computed tomography (SPECT) image pairs. We trained on 40 pairs of images containing all modalities and then tested on 50 pairs of MRI-CT, MRI-PET, and MRI-SPECT images, respectively. All images are aligned and have a size of 256$\times$256 pixels.

\textbf{Metrics.} We quantitatively analyzed the fusion results using six metrics: SD, AG, SF, sum of the correlations of differences (SCD), visual information fidelity for fusion (VIFF), and structural similarity index measure (SSIM). 
AG and SF are key indicators of image texture complexity and edge sharpness. SD reflects the dispersion of pixel intensities. VIFF and SSIM are evaluation metrics that more closely align with the human visual perception system, measuring the extent to which the fused image retains the effective information and structural features of the source images. SCD measures the amount of information transferred from the source images to the fused image. 
Higher metrics indicate better fusion performance. 

\textbf{Comparison methods.} We compare the fusion results with eight SOTA methods: EMMA (2024)~\cite{Zhao_EMMA}, MMDRFuse (2024)~\cite{Deng2024MMDRFuseDM}, MMAE (2024)~\cite{Wang2024MMAEAU}, MLFuse (2025)~\cite{10856398_MLFuse},  VDMUFusion (2025)~\cite{10794610_VDMFusion}, DSAGAN (2021)~\cite{Fu2021DSAGANAG}, FATFusion (2024)~\cite{Tang2024FATFusionAF}, and DM-FNet (2025)~\cite{He2025DMFNetUM}. It is worth noting that DSAGAN, FATFusion, and DM-FNet are applied exclusively to MIF.

\textbf{Implementation details.} The experiments were conducted on a server with an Intel(R) Xeon(R) Silver 4314 CPU @ 2.40GHz and four NVIDIA GeForce RTX 3090 GPUs, using the PyTorch framework. The images are randomly cropped into 64$\times$64 patches with a batch size of 32 during training. We train the network for 1000 epochs using the Adam optimizer with an initial learning rate of 1e-4.

\subsubsection{Ablation experiment}
The ablation study is designed to verify three key aspects: the unsuitability of combining LKC with BN for unsupervised image fusion; the necessity of the hybrid normalization strategy (IN+GN); and the effectiveness of LKC and MPAFM. The results are presented in \Cref{fig:Fig1} and \Cref{tab:xrsy}.

In \Cref{tab:xrsy} (a)(b)(i)(k), the model performs poorly with BN, and its performance degrades further when combined with LKC. This is likely because inter-sample interactions in BN cause data smoothing, which reduces effective features. Consequently, LKC's large receptive field provides limited or even detrimental effects on detail preservation.

In \Cref{tab:xrsy} (b)-(g), LN improves upon BN by mitigating inter-sample interference. BCN is still constrained by BN's smoothing effect, yielding performance between that of LN and BN. GN further enhances performance by considering feature independence. However, using IN throughout the network leads to poor fusion results; while effective at the input stage, its global application can excessively amplify distribution differences between modalities and over-emphasize certain features. The IN+GN configuration significantly outperforms all single-normalization strategies. This demonstrates that IN is crucial at the initial stage for handling the statistical gap between input modalities, while GN provides stable and efficient normalization for deeper, more abstract features.

In \Cref{tab:xrsy} (h)-(k), IN in the InitBlock is indispensable for maintaining image properties, and its combination with LKC significantly enhances performance. MPAFM effectively transmits key features by recalibrating the decoder input. In summary, IN and GN unlock the model's performance potential, while LKC and MPAFM are the key components that help it reach its upper limit.

\subsubsection{Analysis of group size in GN}
\begin{figure}[!t]
\centering
    \includegraphics[width=1.0\linewidth]{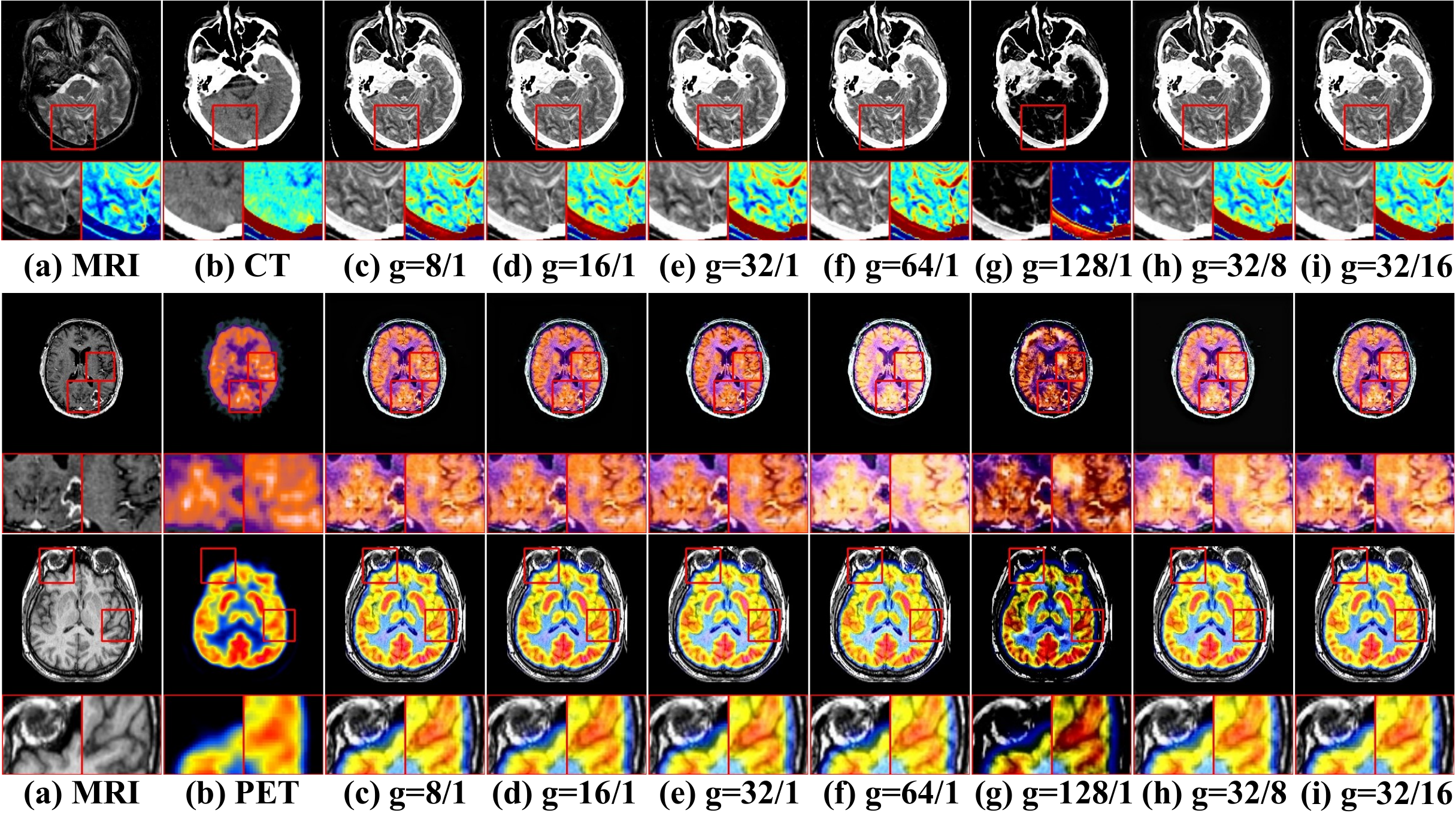}
    \caption{Subjective results of the GN with different group sizes.}
    \label{fig:gn}
\end{figure}

We experimented with the number in different GN (g1 in LKCBlock and g2 in LKDCBlock), as shown in \Cref{fig:gn} and \Cref{tab:GN}. 

In LKDCBlock, where each convolution kernel focuses on a single feature map, a larger group number increases the number of independently normalized feature maps. This can amplify distribution differences, causing certain features to be exaggerated and leading to poor subjective fusion results. This is why we set the GN group number in LKDCBlock to 1. 

In LKCBlock, fusion results are better when the group number is below 32, with the optimal results occurring at 32.
The fusion results become completely unstable when g=c in LKCBlock (i.e., IN). LKC-FUNet uses concatenated inputs, requiring the integration of multi-source feature maps at each step. Using IN throughout the process causes the feature distributions across different image sources to lose consistency, resulting in unstable fusion and potentially introducing artifacts or black spots. This is why we only use IN in the InitBlock.

\begin{table}[!t]
\footnotesize
\setlength\tabcolsep{5pt}
\vspace{-0.5cm}
\caption{Objective results of the GN with different group sizes. Red: best, blue: second-best.}
\label{tab:GN}
\begin{center}
    \begin{tabular}{@{}llllllllllll@{}} \hline
    No. & g1 & g2 & SD & AG & SF & SCD & VIFF & SSIM \\ \hline
    c & 8 & 1 & 78.804 & \textbf{\textcolor{red}{10.616}} & \textbf{\textcolor{red}{38.508}} & 1.383 & 0.547 & 0.717 \\
    d & 16 & 1 & 79.467 & 9.374 & 34.200 & 1.516 & 0.555 & \textbf{\textcolor{red}{0.724}} \\
    e & 32 & 1 & \textbf{\textcolor{blue}{83.422}} & 9.716 & 35.381 & \textbf{\textcolor{red}{1.618}} & \textbf{\textcolor{red}{0.586}} & \textbf{\textcolor{blue}{0.722}} \\
    f & 64 & 1 & \textbf{\textcolor{red}{88.935}} & \textbf{\textcolor{blue}{10.043}} & 36.822 & \textbf{\textcolor{blue}{1.574}} & 0.543 & 0.668 \\
    g & 128 & 1 & 68.829 & 9.017 & \textbf{\textcolor{blue}{37.992}} & 0.439 & 0.407 & 0.624 \\
    h & 32 & 8 & 82.202 & 9.595 & 34.403 & 1.529 & \textbf{\textcolor{blue}{0.559}} & 0.581 \\
    i & 32 & 16 & 81.082 & 9.579 & 35.582 & 1.427 & 0.557 & 0.719 \\ \hline
\vspace{-0.5cm}
\end{tabular}
\end{center}
\end{table}

\subsubsection{Comparison with SOTA methods}
\begin{figure*}[t!]
\centering
   \includegraphics[width=1.0\linewidth]{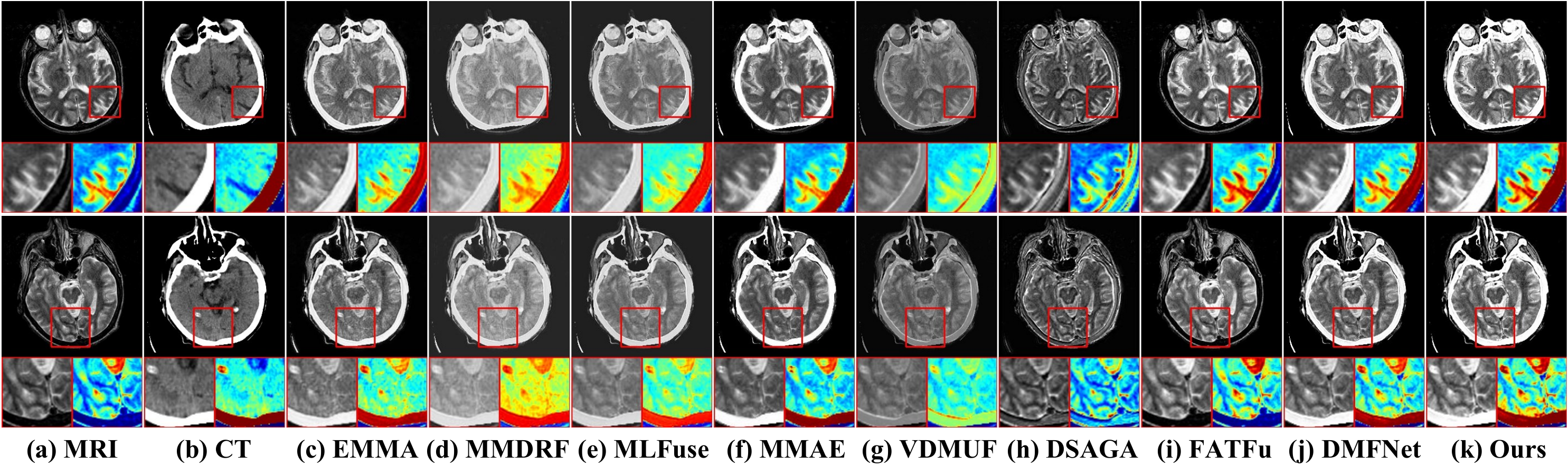}
    \caption{Visual comparison with eight SOTA methods in MRI-CT.
    }\label{fig:mri-ct}
\end{figure*}

\textbf{Qualitative Comparison}. 
\Cref{fig:mri-ct} presents a visual comparison using two representative pairs of MRI and CT. CT images provide high-contrast information for bones and hard tissues, while MRI (T1, T2) reveals signal variations in soft tissues and lesion areas.
EMMA and FATFusion lose crucial information from one modality, while MMDRFuse, MLFuse, and VDMUFusion produce images with low contrast, artifacts, or blurred lesion boundaries. DSAGAN introduces artifacts in some soft tissue areas, affecting lesion visibility. MMAE achieves good overall fusion quality, but the hierarchical sense in some soft tissue areas remains suboptimal. In contrast, our method, alongside DM-FNet, excels in generating sharp bone edges and smooth soft-tissue transitions, providing clearer diagnostic information.

\begin{figure*}[t!]
\centering
   \includegraphics[width=1.0\linewidth]{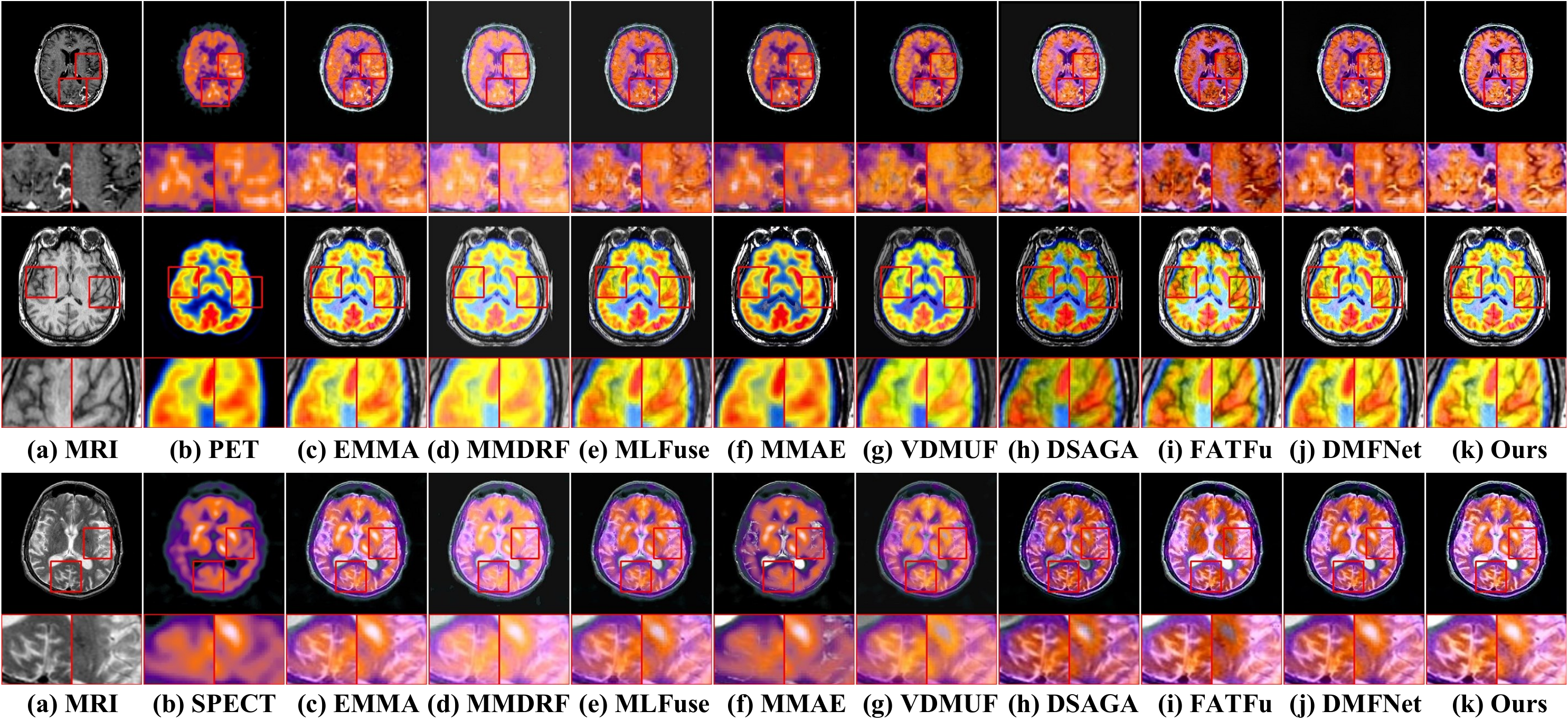}
    \caption{Visual comparison with eight SOTA methods in MRI-PET and MRI-SPECT.
    }\label{fig:mri-fun}
    \vspace{-0.5cm}
\end{figure*}

MRI functional image fusion (\Cref{fig:mri-fun}) requires highlighting metabolic hotspots while preserving anatomical context. EMMA, MMDRFuse, and MMAE allow functional information to obscure anatomical details. VDMUFusion weakens metabolic information and causes severe texture loss. DSAGAN's aggressive contrast enhancement introduces background artifacts. FATFusion preserves MRI structure but completely loses the high-intensity metabolic regions from the functional image. MLFuse also weakens some details. 
In contrast, our method and DM-FNet effectively integrate the functional tracer's intensity and distribution into the precise MRI anatomical background, which is more conducive to accurate clinical diagnosis.

\textbf{Quantitative Comparison}.
\Cref{MIF} displays the objective results for six metrics across three datasets. 
LKC-FUNet achieved the highest or second-highest AG and SF values. This is consistent with the subjective visual performance: the fused images have extremely clear edge contours and rich texture details. 
The highest SD values indicate that our fused images possess higher contrast and a wider dynamic range, making different tissues and lesions more distinguishable. 
Furthermore, LKC-FUNet leads in the VIFF metric and ranks among the top three for SSIM, demonstrating that the fusion results are statistically informative and the highest perceptual quality. 
Finally, the top SCD scores signify that LKC-FUNet minimizes information loss and maximally preserves key information from both modalities.

In summary, our method leverages hybrid normalization to precisely preserve modality-specific information, while integrating the powerful capability of LKC to capture complex structures and textures.
This achieves a fine-grained presentation of anatomical details, sharp tissue boundaries, and faithful preservation of functional tracer distributions, providing clinicians with more reliable diagnostic support.
\begin{table}[!t]
\footnotesize
\setlength\tabcolsep{4pt}
\caption{Quantitative comparisons in MIF. Red: best results, blue: second-best.}
\begin{center}
    \begin{tabular}{@{}lllllll@{}}\hline
    & \multicolumn{6}{l}{\textbf{Dataset: MRI-CT Image Fusion (50 pairs)\footnotemark[1] }} \\
    Methods  & SD & AG & SF & SCD & VIFF & SSIM \\ \hline
    EMMA~\cite{Zhao_EMMA}       & 86.493 &  7.534 & 27.405 & 1.464 & 0.421 & 0.668 \\
    MMDRFuse~\cite{Deng2024MMDRFuseDM} & 75.768 &  5.723 & 22.993 & 1.263 & 0.371 & 0.263 \\
    MLFuse~\cite{10856398_MLFuse}  & 71.231 &  6.202 & 25.932 & 1.018 & 0.374 & 0.284 \\
    MMAE~\cite{Wang2024MMAEAU}   & 87.807 &  7.439 & 33.208 & 1.268 & 0.405 & 0.722 \\
    VDMUFusion~\cite{10794610_VDMFusion} & 63.414 &  5.278 & 20.803 & 0.986 & 0.348 & 0.455 \\
    DSAGAN~\cite{Fu2021DSAGANAG}  & 56.790 &  8.382 & 29.278 & 0.533 & 0.185 & 0.463 \\
    FATFusion~\cite{Tang2024FATFusionAF} & 72.680 &  7.688 & 26.901 & 0.906 & 0.119 & 0.717 \\
    DM\_FNet~\cite{He2025DMFNetUM}  & \textbf{\textcolor{blue}{89.708}} & \textbf{\textcolor{blue}{9.000}} & \textbf{\textcolor{red}{38.759}} & \textbf{\textcolor{blue}{1.440}} & \textbf{\textcolor{blue}{0.436}} & \textbf{\textcolor{red}{0.742}} \\
    Ours & \textbf{\textcolor{red}{93.483}} & \textbf{\textcolor{red}{9.190}} & \textbf{\textcolor{blue}{37.203}} & \textbf{\textcolor{red}{1.561}} & \textbf{\textcolor{red}{0.458}} & \textbf{\textcolor{blue}{0.727}} \\ \hline

    & \multicolumn{6}{l}{\textbf{Dataset: MRI-PET Image Fusion (50 pairs)\footnotemark[1] }} \\
    Methods  & SD & AG & SF & SCD & VIFF & SSIM \\ \hline
    EMMA~\cite{Zhao_EMMA}        & \textbf{\textcolor{blue}{74.011}} &  7.151 & 22.547 & 1.602 & 0.555 & 0.668 \\
    MMDRFuse~\cite{Deng2024MMDRFuseDM}    & 71.833 &  5.723 & 20.099 & \textbf{\textcolor{blue}{1.604}} & 0.539 & 0.262 \\
    MLFuse~\cite{10856398_MLFuse}      & 63.569 &  6.439 & 21.971 & 1.360 & 0.518 & 0.327 \\
    MMAE~\cite{Wang2024MMAEAU}        & 68.674 &  7.207 & 29.287 & 1.189 & 0.431 & 0.717 \\
    VDMUFusion~\cite{10794610_VDMFusion}  & 59.793 &  5.015 & 16.203 & 1.204 & 0.448 & 0.434 \\
    DSAGAN~\cite{Fu2021DSAGANAG}      & 59.380 &  7.686 & 25.612 & 1.097 & 0.408 & 0.465 \\
    FATFusion~\cite{Tang2024FATFusionAF}   & 69.284 & \textbf{\textcolor{blue}{8.760}} & 30.089 & 1.278 & 0.402 & \textbf{\textcolor{red}{0.727}} \\
    DM\_FNet~\cite{He2025DMFNetUM}    & 73.751 &  8.698 & 29.258 & 1.570 & \textbf{\textcolor{blue}{0.581}} & 0.693 \\
    Ours        & \textbf{\textcolor{red}{80.593}} & \textbf{\textcolor{red}{10.277}} & \textbf{\textcolor{red}{35.394}} & \textbf{\textcolor{red}{1.688}} & \textbf{\textcolor{red}{0.625}} & \textbf{\textcolor{blue}{0.724}} \\ \hline
    
    & \multicolumn{6}{l}{\textbf{Dataset: MRI-SPECT Image Fusion (50 pairs)\footnotemark[1] }} \\
    Methods  & SD & AG & SF & SCD & VIFF & SSIM \\ \hline
    EMMA~\cite{Zhao_EMMA}        & 68.472 &  6.378 & 20.555 & 1.528 & 0.606 & 0.664 \\
    MMDRFuse~\cite{Deng2024MMDRFuseDM}    & 68.655 &  5.499 & 18.248 & \textbf{\textcolor{blue}{1.597}} & 0.589 & 0.263 \\
    MLFuse~\cite{10856398_MLFuse}      & 59.391 &  5.686 & 19.761 & 1.204 & 0.556 & 0.354 \\
    MMAE~\cite{Zhao_EMMA}        & 57.345 &  5.780 & 24.343 & 0.707 & 0.389 & 0.708 \\
    VDMUFusion~\cite{10794610_VDMFusion}  & 50.915 &  4.160 & 13.564 & 1.078 & 0.438 & 0.413 \\
    DSAGAN~\cite{Fu2021DSAGANAG}      & 58.170 &  7.396 & 24.569 & 1.104 & 0.513 & 0.521 \\
    FATFusion~\cite{Tang2024FATFusionAF}   & \textbf{\textcolor{blue}{73.948}} & \textbf{\textcolor{blue}{7.988}} & \textbf{\textcolor{blue}{27.301}} & 1.431 & 0.561 & 0.711 \\
    DM\_FNet~\cite{He2025DMFNetUM}    & 68.524 &  7.769 & 26.713 & 1.446 & \textbf{\textcolor{blue}{0.620}} & \textbf{\textcolor{red}{0.721}} \\
    Ours        & \textbf{\textcolor{red}{76.190}} & \textbf{\textcolor{red}{9.681}} & \textbf{\textcolor{red}{33.546}} & \textbf{\textcolor{red}{1.606}} & \textbf{\textcolor{red}{0.676}} & \textbf{\textcolor{blue}{0.713}} \\ \hline
    \end{tabular}
\label{MIF}
\end{center}
\vspace{-0.8cm}
\end{table}

\subsection{Infrared and visible image fusion}
\subsubsection{Setup}
\textbf{Datasets.} We perform comparative experiments on three mainstream datasets: MSRS \cite{Tang2022PIAFusionAP}, M$^3$FD\cite{liu2022target} and TNO\cite{Toet2017TheTM_tno}. We trained the model on 1083 pairs of images from MSRS and then evaluated the performance on 361 pairs of test data. In addition, the models are applied to the M$^3$FD and TNO datasets without fine-tuning to verify the generalization performance.

\textbf{SOTA methods.} We compare the fusion results with nine SOTA methods: EMMA (2024)~\cite{Zhao_EMMA}, MMDRFuse (2024)~\cite{Deng2024MMDRFuseDM}, MLFuse (2025)~\cite{10856398_MLFuse}, MMAE (2024)~\cite{Wang2024MMAEAU}, VDMUFusion (2025)~\cite{10794610_VDMFusion}, LRRFNet (2023)~\cite{Li2023LRRNetAN}, DCINN (2024)~\cite{wang2024general}, SHIP (2024)~\cite{10655996}, and MaeFuse (2024)~\cite{Li2024MaeFuseTO}. It is worth noting that the latter four methods are applied only to IVIF. The training strategy and evaluation metrics are the same as for MIF.

\subsubsection{Comparison with SOTA methods}
\begin{figure*}[!t]
\centering
    \includegraphics[width=0.85\linewidth]{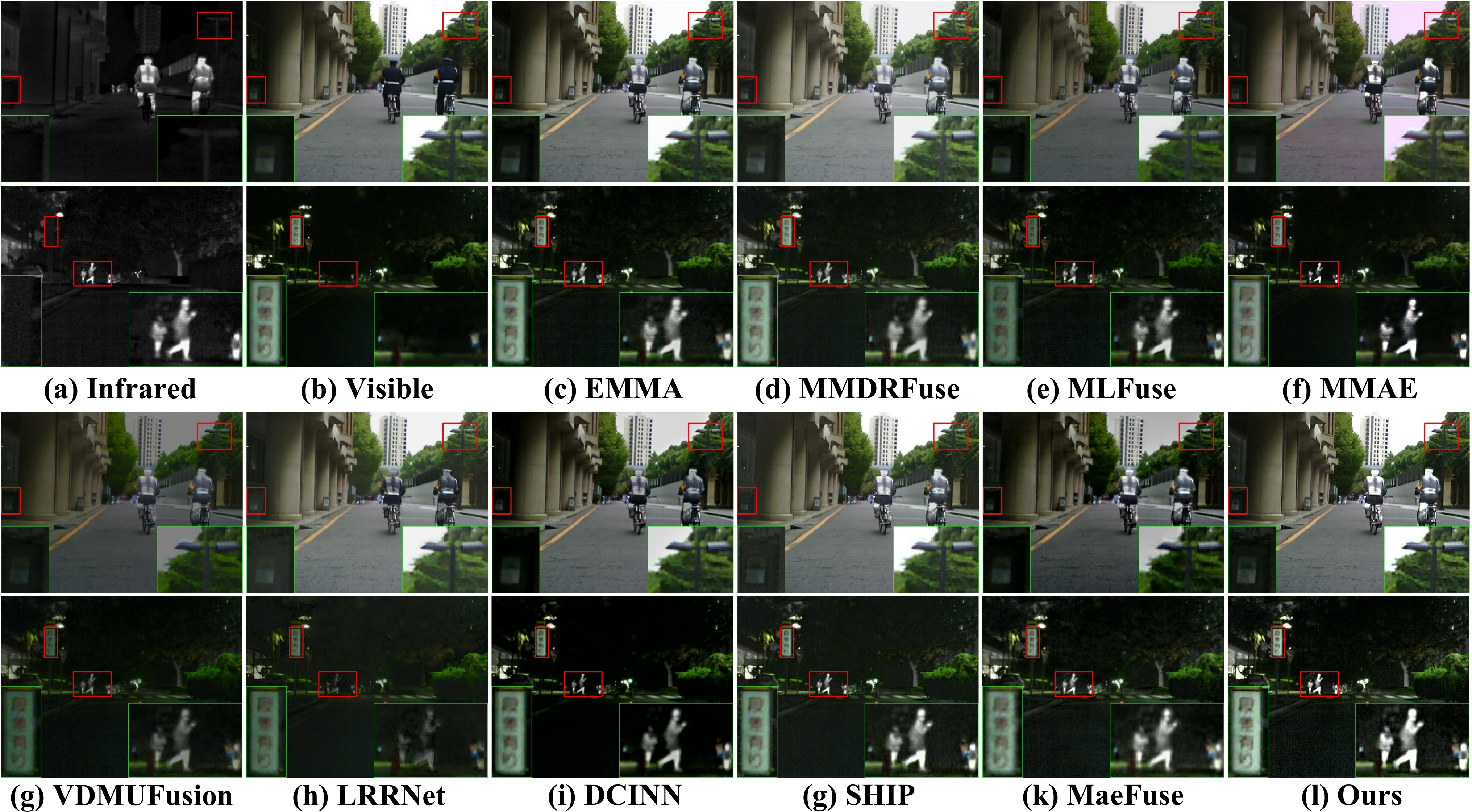}
   \caption{Visual comparison with nine SOAT methods on MSRS in IVIF.
   }\label{fig:MSRS}
\end{figure*}

\begin{figure*}[!t]
\centering
    \includegraphics[width=0.85\linewidth]{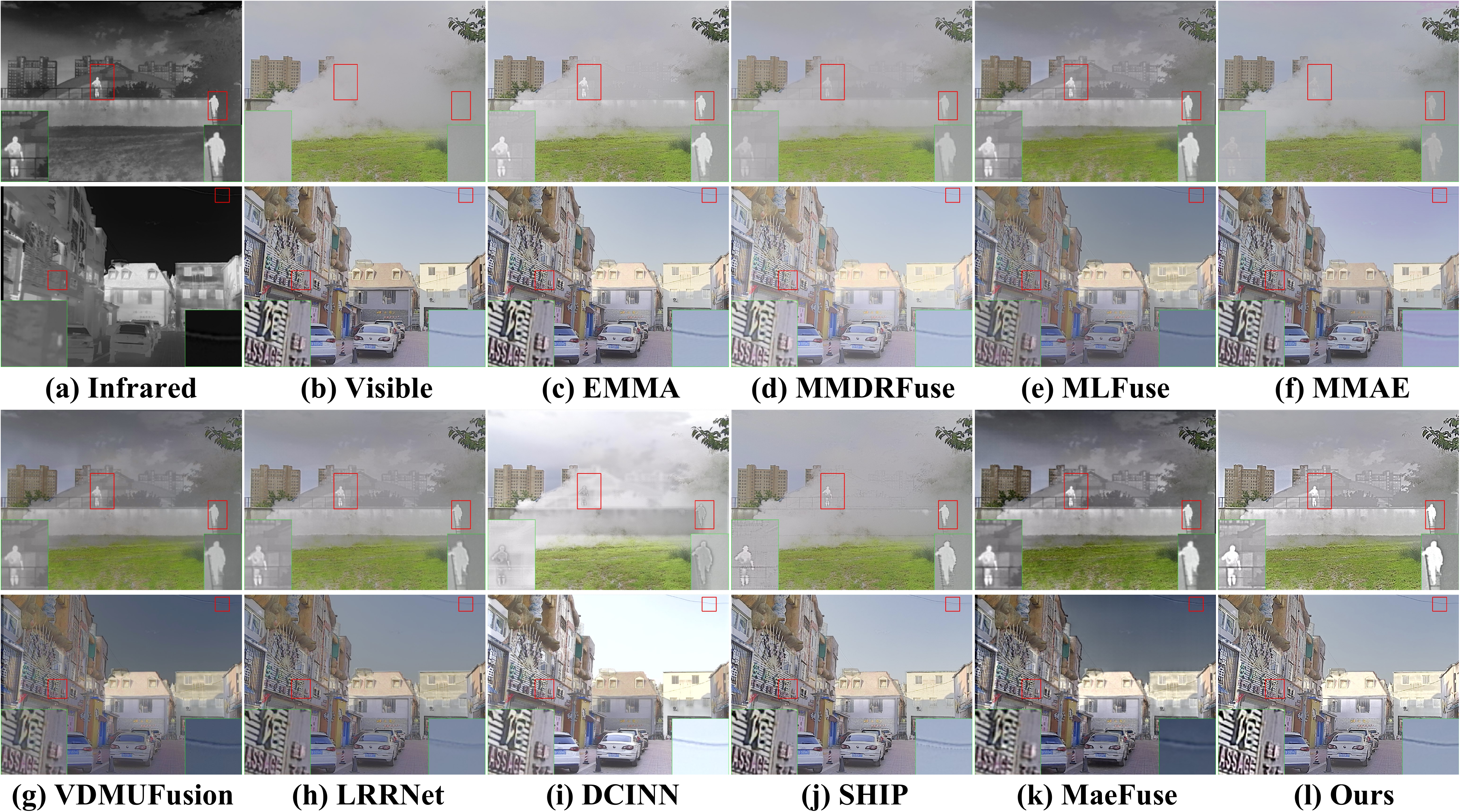}
   \caption{Visual comparison with nine SOAT methods on M$^3$FD in IVIF.
   }\label{fig:M3FD}
\end{figure*}

\begin{figure*}[!t]
\centering
    \includegraphics[width=0.85\linewidth]{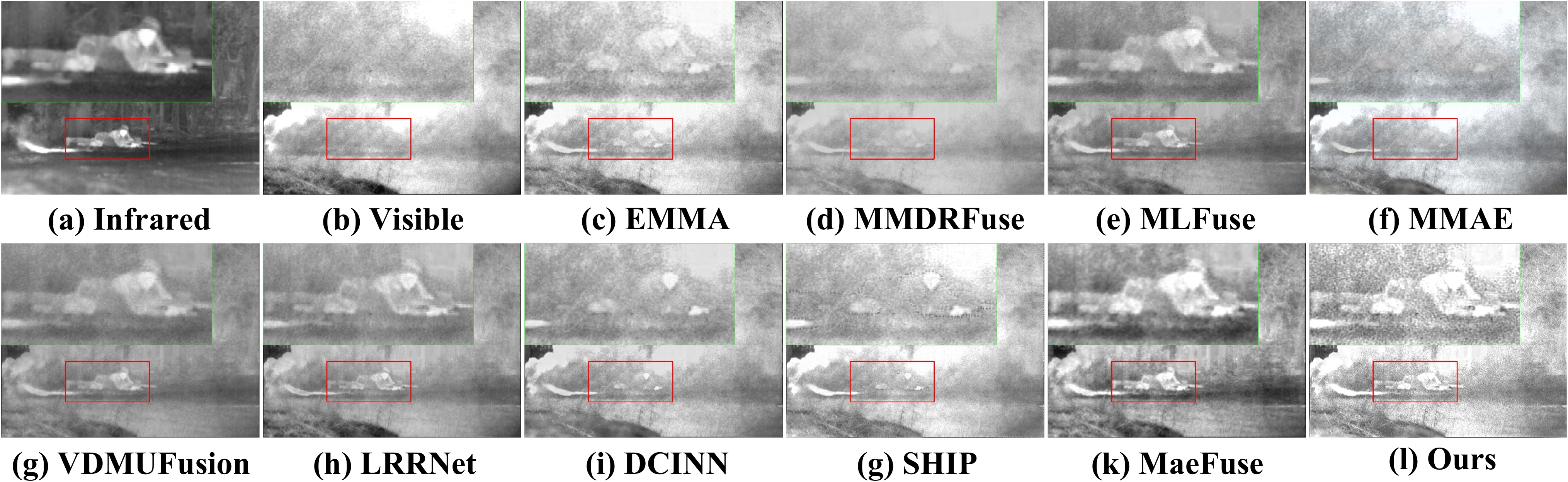}
   \caption{Visual comparison with nine SOAT methods on TNO in IVIF.
   }\label{fig:TNO}
\end{figure*}

\textbf{Qualitative Comparison.} As shown in \cref{fig:MSRS}\cref{fig:M3FD} and \cref{fig:TNO}, our method demonstrates strong robustness across various adverse environments.
In the MSRS dataset (\Cref{fig:MSRS}), VDMUFusion and LRRNet exhibit poor object recovery in dark regions. In contrast, our method highlights targets in low light, preserving the background details from the visible image while injecting the target saliency from the infrared image. 

In the M³FD dataset (\Cref{fig:M3FD}), where methods like EMMA, MMDRFuse, MMAE, DCINN, and SHIP struggle to recover key information in smoky scenes, our method successfully preserves important details within the smoke. It also avoids the distortion or artifacts seen in the results of VDMUFusion and MaeFuse when processing background areas like the sky. 

In the TNO dataset (\Cref{fig:TNO}), when MMDRFuse and MMAE completely lose the target in the smoke and EMMA, DCINN, and SHIP fail to preserve it clearly, our method still generates sharp, discernible target contours. 

In summary, our method highlights objects in dark areas, and scenes and targets in smoke are equally visible, demonstrating strong robustness in adverse environments. This unique capability enhances our understanding of the depicted scenes.

\begin{table}[!tbp]
\footnotesize
\setlength\tabcolsep{4pt}
\caption{Quantitative comparisons in IVIF. Red: best, blue: second-best.}
\begin{center}
    \begin{tabular}{@{}lllllll@{}}\hline
        \multicolumn{7}{c}{\textbf{Dataset: MSRS (361 pairs of IR-VIS images) \cite{Tang2022PIAFusionAP}}} \\
        Methods  & SD & AG & SF & SCD & VIFF & SSIM \\ \hline
        EMMA~\cite{Zhao_EMMA}       & {\bfseries\textcolor{blue}{44.578}} & 3.785 & 11.544 & 1.629 & 0.773 & 0.712 \\
        MMDRFuse~\cite{Deng2024MMDRFuseDM}   & 40.248 & 3.494 & 10.426 & 1.595 & 0.729 & 0.691 \\
        MLFuse~\cite{10856398_MLFuse}     & 33.879 & 2.930 & 8.803  & 1.519 & 0.591 & 0.705 \\
        MMAE~\cite{Wang2024MMAEAU}       & 40.787 & 3.189 & 10.490 & 1.536 & 0.636 & {\bfseries\textcolor{blue}{0.714}} \\
        VDMUFu~\cite{10794610_VDMFusion} & 23.597 & 2.255 & 6.544  & 1.267 & 0.389 & {\bfseries\textcolor{red}{0.721}} \\
        LRRNet~\cite{Li2023LRRNetAN}     & 31.757 & 2.651 & 8.464  & 0.791 & 0.414 & 0.614 \\
        DCINN~\cite{wang2024general}      & 40.112 & 3.334 & 10.467 & 1.478 & 0.706 & 0.583 \\
        SHIP~\cite{10655996}       & 41.131 & {\bfseries\textcolor{blue}{3.933}} & {\bfseries\textcolor{blue}{11.803}} & 1.512 & 0.701 & 0.665 \\
        MaeFuse~\cite{Li2024MaeFuseTO}    & 38.297 & 3.462 & 9.560  & {\bfseries\textcolor{red}{1.712}} & {\bfseries\textcolor{blue}{0.796}} & 0.704 \\
        Ours       & {\bfseries\textcolor{red}{46.574}} & {\bfseries\textcolor{red}{4.939}} & {\bfseries\textcolor{red}{15.377}} & {\bfseries\textcolor{blue}{1.682}} & {\bfseries\textcolor{red}{0.883}} & 0.680 \\ \hline

        \multicolumn{7}{c}{\textbf{Dataset: M$^3$FD (300 pairs of IR-VIS images) \cite{liu2022target}}} \\
       Methods  & SD & AG & SF & SCD & VIFF & SSIM \\ \hline
       EMMA~\cite{Zhao_EMMA}       & 38.251 & 5.340 & 15.216 & 1.494 & 0.491 & 0.702 \\
        MMDRFuse~\cite{Deng2024MMDRFuseDM}   & 26.594 & 3.431 & 10.211 & 1.471 & 0.305 & 0.718 \\
        MLFuse~\cite{10856398_MLFuse}     & 28.298 & 3.288 & 9.779  & 1.567 & 0.406 & {\bfseries\textcolor{red}{0.747}} \\
        MMAE~\cite{Wang2024MMAEAU}       & 31.954 & 4.074 & 12.233 & 1.275 & 0.288 & 0.714 \\
        VDMUFu~\cite{10794610_VDMFusion} & 26.558 & 2.790 & 7.919  & 1.530 & 0.328 & {\bfseries\textcolor{blue}{0.745}} \\
        LRRNet~\cite{Li2023LRRNetAN}     & 27.179 & 3.600 & 10.678 & 1.463 & 0.363 & 0.725 \\
        DCINN~\cite{wang2024general}      & {\bfseries\textcolor{red}{44.020}} & 4.303 & 12.942 & 0.346 & 0.319 & 0.683 \\
        SHIP~\cite{10655996}       & 35.222 & {\bfseries\textcolor{blue}{5.177}} & {\bfseries\textcolor{blue}{15.247}} & 1.311 & 0.385 & 0.693 \\
        MaeFuse~\cite{Li2024MaeFuseTO}    & 36.102 & 3.758 & 9.523  & {\bfseries\textcolor{blue}{1.750}} & {\bfseries\textcolor{blue}{0.609}} & 0.718 \\
        Ours       & {\bfseries\textcolor{blue}{40.235}} & {\bfseries\textcolor{red}{7.414}} & {\bfseries\textcolor{red}{22.102}} & {\bfseries\textcolor{blue}{1.614}} & {\bfseries\textcolor{red}{0.671}} & 0.656 \\ \hline
        
        \multicolumn{7}{c}{\textbf{Dataset: TNO (37 pairs of IR-VIS images) \cite{Toet2017TheTM_tno}}} \\
        Methods  & SD & AG & SF & SCD & VIFF & SSIM \\ \hline
        EMMA~\cite{Zhao_EMMA}       & {\bfseries\textcolor{red}{48.575}} & 5.171 & 12.646 & 1.668 & 0.579 & 0.661 \\
        MMDRFuse~\cite{Deng2024MMDRFuseDM}   & 28.964 & 3.037 & 7.983  & 1.418 & 0.261 & 0.724 \\
        MLFuse~\cite{10856398_MLFuse}     & 36.354 & 3.849 & 10.575 & 1.700 & 0.534 & 0.719 \\
        MMAE~\cite{Wang2024MMAEAU}       & 42.346 & 3.972 & 11.034 & 1.463 & 0.304 & 0.703 \\
        VDMUFu~\cite{10794610_VDMFusion} & 26.634 & 2.787 & 7.378  & 1.522 & 0.308 & {\bfseries\textcolor{red}{0.743}} \\
        LRRNet~\cite{Li2023LRRNetAN}     & 42.224 & 4.104 & 10.438 & 1.507 & 0.415 & 0.653 \\
        DCINN~\cite{wang2024general}      & 35.833 & 3.942 & 10.433 & 1.573 & 0.347 & {\bfseries\textcolor{blue}{0.725}} \\
        SHIP~\cite{10655996}       & 40.149 & {\bfseries\textcolor{blue}{5.004}} & {\bfseries\textcolor{blue}{13.224}} & 1.455 & 0.355 & 0.676 \\
        MaeFuse~\cite{Li2024MaeFuseTO}    & 37.961 & 4.027 & 8.930  & {\bfseries\textcolor{red}{1.790}} & {\bfseries\textcolor{red}{0.626}} & 0.707 \\
        Ours       & {\bfseries\textcolor{blue}{43.800}} & {\bfseries\textcolor{red}{6.631}} & {\bfseries\textcolor{red}{16.713}} & {\bfseries\textcolor{blue}{1.670}} & {\bfseries\textcolor{blue}{0.608}} & 0.636 \\ \hline
    \end{tabular}
\label{tab:VIF}
\end{center}
\vspace{-0.7cm}
\end{table}
\textbf{Quantitative Comparison}. 
Quantitative results are shown in \Cref{tab:VIF}. Our approach achieves SOTA performance on most key metrics across three datasets, confirming its applicability to various environmental conditions and object classes. The key metrics of AG and SF demonstrate our method's ability to preserve edge contours and texture details. The excellent performance on SD, SCD, and VIFF also indicates that the fused images have higher contrast and richer visual information, generating more informative images that better align with human visual perception. All experimental results show that LKC-FUNet can simultaneously preserve the integrity of source image features and generate information-rich fused images.

\subsubsection{Downstream IVIF applications}
\begin{table}[!t]
\footnotesize
\setlength\tabcolsep{2pt}
\caption{IoU(\%) for semantic segmentati on MSRS dataset. Red: best, blue: second-best.}
\begin{center}
    \begin{tabular}{@{}lllllllllll@{}}\hline
    Methods  & Unl & Car & Per & Bik & Cur & CS & GD & CC & Bu & mIoU \\ \hline
    IR & 97.1 & 83.0 & 66.3 & 49.1 & 34.0 & 50.9 & 59.8 & 40.4 & 51.9 & 59.2 \\
    VI & 97.4 & 85.7 & 58.3 & 55.8 & 40.2 & 58.4 & 65.2 & 50.0 & 52.6 & 62.6 \\
    EMMA~\cite{Zhao_EMMA} & {97.7} & 86.1 & 67.8 & 57.6 & \textbf{\textcolor{blue}{46.6}} & 61.8 & 66.9 & \textbf{\textcolor{red}{55.6}} & 56.3 & 66.3 \\
    MMDRF~\cite{Deng2024MMDRFuseDM} & {97.6} & {86.1} & \textbf{\textcolor{blue}{68.0}} & \textbf{\textcolor{red}{58.2}} & 44.3 & \textbf{\textcolor{red}{63.7}} & 69.0 & 53.8 & 60.0 & \textbf{\textcolor{blue}{66.8}} \\
    MLFuse~\cite{10856398_MLFuse} & {97.6} & 85.6 & 67.1 & 55.8 & 42.4 & 63.2 & {67.3} & 55.3 & 56.6 & 65.7 \\
    MMAE~\cite{Wang2024MMAEAU} & {97.7} & \textbf{\textcolor{red}{87.2}} & 66.7 & \textbf{\textcolor{blue}{58.2}} & 42.9 & 60.6 & 64.7 & 53.4 & 56.7 & 65.3 \\
    VDMUF~\cite{10794610_VDMFusion} & 97.5 & 84.4 & 66.3 & 57.9 & 44.3 & 60.7 & 66.8 & {54.9} & 56.5 & 65.5 \\
    LRRNet~\cite{Li2023LRRNetAN} & {97.6} & 85.2 & 66.9 & 55.5 & 44.6 & 61.8 & 64.7 & 51.3 & \textbf{\textcolor{red}{61.0}} & 65.4 \\
    DCINN~\cite{wang2024general} & \textbf{\textcolor{blue}{97.7}} & \textbf{\textcolor{blue}{86.6}} & 67.3 & 57.0 & 42.0 & 59.0 & \textbf{\textcolor{blue}{67.3}} & 54.2 & 58.8 & 65.5 \\
    SHIP~\cite{10655996} & 97.6 & 85.8 & 67.3 & 56.0 & 45.9 & 62.6 & 64.6 & 52.4 & 54.6 & 65.2 \\
    MaeFuse~\cite{Li2024MaeFuseTO} & {97.6} & 86.0 & 66.7 & 57.3 & 44.9 & 62.3 & 68.9 & 52.7 & 56.5 & 65.9 \\
    Ours & \textbf{\textcolor{red}{97.7}} & {86.2} & \textbf{\textcolor{red}{68.9}} & {58.1} & \textbf{\textcolor{red}{46.1}} & \textbf{\textcolor{blue}{63.3}} & \textbf{\textcolor{red}{69.1}} & \textbf{\textcolor{blue}{55.5}} & \textbf{\textcolor{blue}{60.4}} & \textbf{\textcolor{red}{67.3}} \\ \hline
    \end{tabular}
\label{Seg}
\end{center}
\vspace{-0.7cm}
\end{table}

\begin{figure*}[t!]
\centering
    \includegraphics[width=0.95\linewidth]{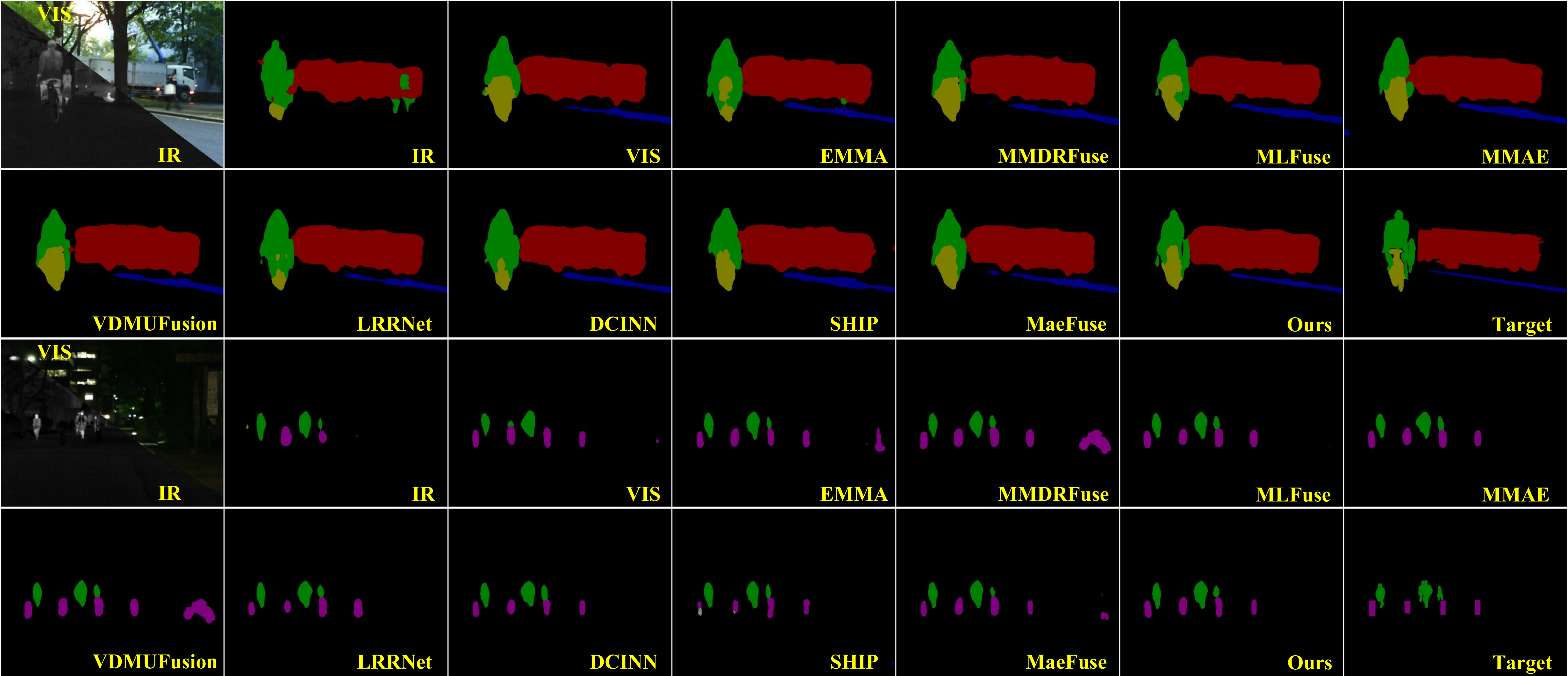}
   \caption{The results of the qualitative analysis of the segmentation tasks, from top to bottom, every two rows form a comparison group, respectively.
   }\label{fig:Seg}
   \vspace{-0.7cm}
\end{figure*}

This section applies the fused images from IR, VIS, and SOTA methods to multimodal semantic segmentation to investigate the benefits of information fusion for downstream tasks. We perform semantic segmentation on the MSRS dataset~\cite{Tang2022PIAFusionAP}, with the data split following~\cite{Tang2022PIAFusionAP}. The nine labels are: background, bump, color cone, guardrail, curve, bicycle, person, car parking, and car. We choose DeeplabV3 as the backbone and Intersection over Union (IoU) as the evaluation metric. The epoch, batch size, optimizer, and initial learning rate for training were set to 500, 4, an SGD optimizer, and 1e-3, respectively. The segmentation results are shown in \Cref{Seg} and \Cref{fig:Seg}. LKC-FUNet better integrates the edge and contour information from the source images, enhancing the model's ability to perceive object boundaries and avoiding segmentation class errors caused by lost targets.

\subsection{Statistical significance analysis}
We conducted pairwise comparisons between LKC-FUNet and each baseline method on a per-image basis. For each metric, we computed difference vectors and applied the Shapiro-Wilk test to assess normality. When paired differences followed a normal distribution (p$>$0.05), we used the paired t-test; otherwise, the Wilcoxon signed-rank test was employed. To control the family-wise error rate across multiple comparisons, all raw p-values were adjusted using the Holm-Bonferroni method. An adjusted p-value below 0.05 was considered statistically significant. 
 
\begin{figure*}[t!]
\centering
   \includegraphics[width=1.0\linewidth]{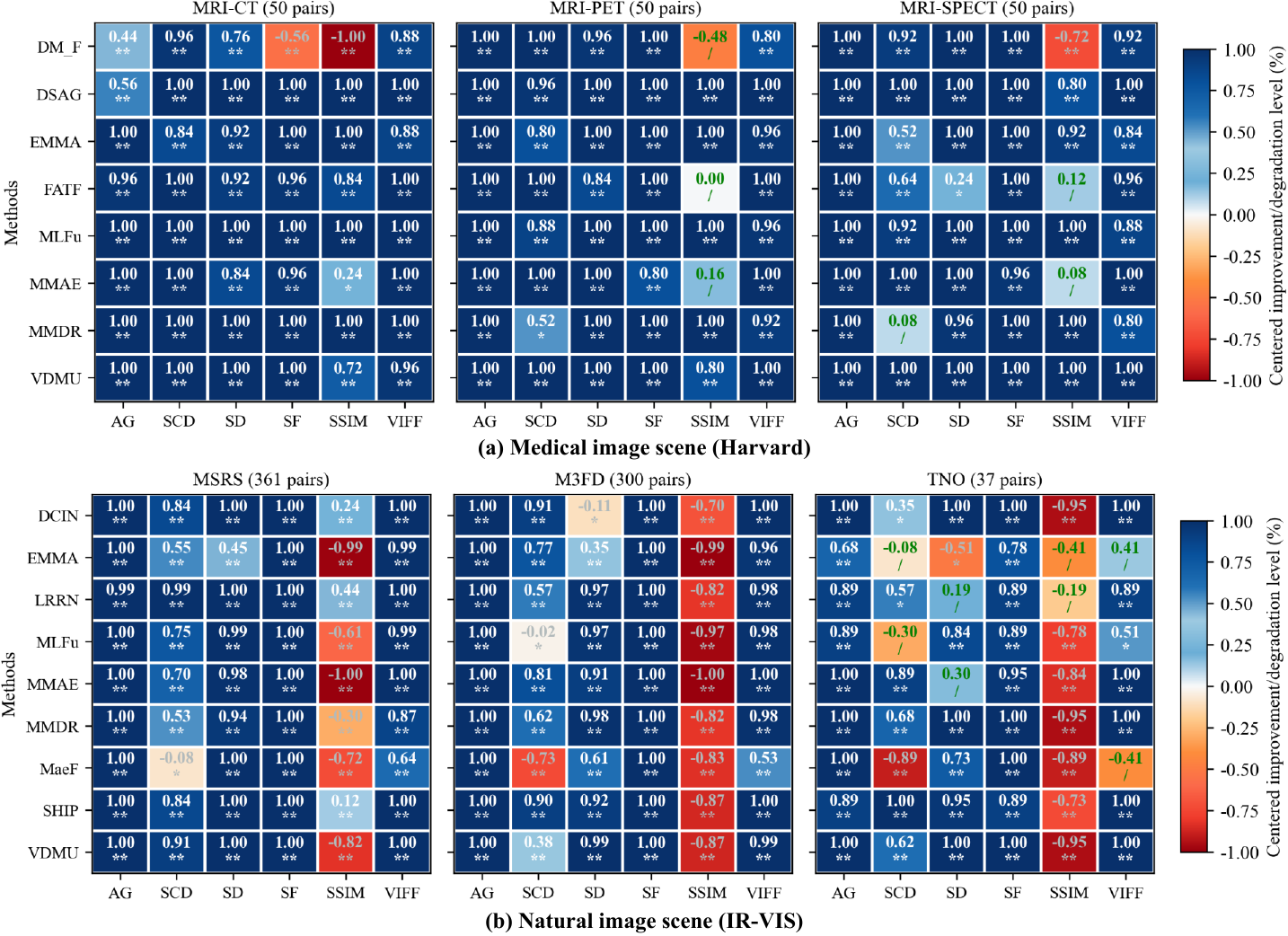}
    \caption{Significance heatmap of quantitative results between LKC-FUNet and SOTA methods. Each cell represents the centered improvement rate. Color and number indicate improvement $(0,1)$/\textcolor{gray}{degradation} $(-1,0)$ level. For instance, 0.44 and -0.56 correspond to improvements in 72\% and 22\% (\textcolor{gray}{a degradation}) of samples, respectively. Symbols denote statistical confidence: \textbf{{non-significant (p$>$0.05, ``/'')}}, significant (p$<$0.05, ``*''), and highly significant (p$<$0.001, ``**'').
    }\label{fig:Statistical}
    \vspace{-0.5cm}
\end{figure*}

To visualize the overall trend, we constructed a significance heatmap, as shown in \Cref{fig:Statistical}. $r$ denotes the improvement rate across all samples, each cell in the heatmap represents the centered improvement rate $r_c=(r-0.5)\times2\in[-1,1]$. This metric intuitively indicates whether the majority of samples improved in the positive direction. For example, 0.44 indicates improvement in 72\% of samples, while -0.56 indicates improvement in only 22\% (a degradation). The symbols denote the level of statistical confidence after Holm's correction: {non-significant (p$>$0.05, ``/'')}, significant (p$<$0.05, ``*''), and highly significant (p$<$0.001, ``**'').

In MIF, LKC-FUNet demonstrates statistically significant and directionally consistent improvements over the best-performing baselines in SD, SF, AG, and VIFF, with $r$ values approaching 100\%. Although SSIM and SCD exhibit weaker significance, several baselines maintain consistent improvement trends. In IVIF, although perceptual structural (SSIM) gain is limited, LKC-FUNet achieves statistically reliable fusion quality enhancement in contrast and detail metrics.

\subsection{Fusion Efficiency}
This section tests the parameter count, FLOPs, and inference time of various SOTA methods on a 256$\times$256 (MIF) resolution. Due to differences in data processing and testing strategies among methods, we use the DeepSpeed library to uniformly calculate the time it takes for a model to produce a fused result, ensuring a fair measurement. The Ranking reflects the total rank across all metrics on the three MIF datasets to comprehensively analyze fusion performance and computational efficiency.

\begin{table}[!t]
\footnotesize
\setlength\tabcolsep{4pt} 
\caption{Model efficiency and performance comparison on MIF ($256\times256$). CTime (ms) and GTime (ms): CPU/GPU inference time. Ranking: overall ranking (three medical datasets). Boldface: best results.}
\label{tab:efficiency_transposed}
\begin{center}
\begin{tabular}{l|lllll}\hline
Methods & Params (K) & FLOPs (G) & GTime & CTime  &Ranking \\ \hline
EMMA~\cite{Zhao_EMMA} & 1520 & 17 & 34 & 325 & 24/25/27 \\
MMDRF~\cite{Deng2024MMDRFuseDM} & \textbf{0.113} & \textbf{0.01} & \textbf{0.4} & \textbf{3} & 41/45/34 \\
MLFuse~\cite{10856398_MLFuse} & {112} & {13} & {6} & 57 & 47/46/50 \\
MMAE~\cite{Wang2024MMAEAU} & 842 & 85 & 22 & 114 & 39/41/35\\
VDMUF~\cite{10794610_VDMFusion} & 31770 & 253800 & 8190 & 62150 & 47/46/50\\
DASGAN~\cite{Fu2021DSAGANAG} & 641 & 84 & 10 & 144 & 39/41/35 \\
FATFu~\cite{Tang2024FATFusionAF} & 7770 & 811 & 134 & 1710 & 37/25/19 \\
DM-FNet~\cite{He2025DMFNetUM} & 23774 & 252 & 185 & 6895 &  {11}/20/17\\
Ours & 41170 & 4050 & 230 & 3210 &  \textbf{8}/\textbf{7}/\textbf{7}\\ \hline
\end{tabular}
\end{center}
\label{tab:MIF_time}
\vspace{-0.5cm}
\end{table}
As shown in \Cref{tab:MIF_time}, different methods exhibit a clear trade-off between performance and efficiency. 
Specificially, VDMUFusion, a diffusion model, requires an iterative denoising process, resulting in an inference time of several seconds per image, making it unsuitable for real-time applications. 
EMMA, MMDRFuse, MLFuse, MMAE, and DASGAN have better fusion efficiency but severely sacrifice fusion quality. 
In contrast, our method strikes a superior balance. Although our GPU inference time is slightly higher than that of DM-FNet, our overall performance ranking is consistently first. On a CPU, our inference time is 3210 ms, more than twice as fast as the similarly performing DM-FNet (6895 ms). This result highlights our architecture's stronger potential for deployment in non-GPU environments, such as edge devices.

Notably, LKC introduces higher computational complexity and parameter counts compared to lightweight models, which is a key bottleneck for deployment on resource-constrained devices. However, LKC-FUNet's SOTA performance makes it an excellent candidate for model optimization techniques. Future work will focus on striking a new balance between performance and speed through methods such as knowledge distillation, where LKC-FUNet can serve as a powerful teacher, and network pruning to reduce complexity. These efforts will enable the efficient deployment of our architecture on edge devices.

\section{Conclusion}
\label{sec:conc}
This paper reveals the limitations of batch normalization in MIF tasks, where BN's cross-sample smoothing operation destroys important sparse features in images, which is a deep-seated reason why many models lose details in MIF. Based on this, we proposed LKC-FUNet, which successfully unleashes the potential of large kernel convolution for detail preservation through a hybrid IN+GN normalization strategy and optimizes the feature flow with the MPAFM module. Extensive experiments demonstrate that our method achieves optimal performance in both visual and objective evaluation metrics and can promote the performance of downstream tasks. Although LKC-FUNet has achieved significant success, LKC also brings higher computational complexity and parameter counts, which may become a bottleneck for deploying the model on resource-constrained devices. Therefore, future work could consider other efficient structures for expanding the receptive field, such as dilated convolution and Mamba models.
\section*{Acknowledgments}
This work was supported by the National Natural Science Foundation of China (Nos. 62331008, 62027827, 62221005, and 62276040), the Natural Science Foundation of Chongqing (Nos. {CSTB2024NSCQ-MSX0726}, 2023NSCQ-LZX0047 and CSTB2022NSCQ-MSX0436), {the Science and Technology Research Program of Chongqing Municipal Education Commission (Nos. KJQN202400507 and KJQN202500621), the Chongqing Normal University Foundation Project (No. 23XLB028)}, the Special Grant for Chongqing Postdoctoral Research Program (No. 2024CQBSHTB3046).
{
    \small
    \bibliographystyle{ieeenat_fullname}
    \bibliography{refs}
}

\end{document}